\documentclass[journal]{vgtc}                

\ifpdf
  \pdfoutput=1\relax                   
  \usepackage{graphicx}                
  \DeclareGraphicsExtensions{.pdf,.png,.jpg,.jpeg} 
\else
  \usepackage{graphicx}                
  \DeclareGraphicsExtensions{.eps}     
\fi%

\graphicspath{{figures/}{pictures/}{images/}{./}} 

\usepackage{microtype}                 
\PassOptionsToPackage{warn}{textcomp}  
\usepackage{textcomp}                  
\usepackage{mathptmx}                  
\usepackage{times}                     
\usepackage{cite}                      
\usepackage{tabu}                      
\usepackage{booktabs}                  

\onlineid{0}

\vgtccategory{Research}
\vgtcpapertype{please specify}

\title{A Projector-Camera System Using Hybrid Pixels with Projection and Capturing Capabilities}

\author{Kenta Yamamoto, Daisuke Iwai, and Kosuke Sato}
\authorfooter{
\item
 Kenta Yamamoto, Daisuke Iwai, and Kosuke Sato are with Graduate School of Engineering Science, Osaka University
}

\shortauthortitle{}

\abstract{We propose a novel projector-camera system (ProCams) in which each pixel has both projection and capturing capabilities. Our proposed ProCams solves the difficulty of obtaining precise pixel correspondence between the projector and the camera. We implemented a proof-of-concept ProCams prototype and demonstrated its applicability to a dynamic projection mapping.%
} 


\renewcommand{\manuscriptnotetxt}{}

\nocopyrightspace

\vgtcinsertpkg

\begin{document}


\firstsection{Introduction}

\maketitle

Projection mapping (PM) is a powerful technique for applications such as entertainment~\cite{6193074} and medicine~\cite{00000658-201806000-00024}. Projection images can change the appearance of real objects, and multiple people can see them simultaneously. In recent years, dynamic projection mapping (DPM) systems that stick images onto moving objects have been widely investigated since DPM can improve the immersion of PM~\cite{7223330,https://doi.org/10.1111/cgf.13128}.

In DPM, it is necessary to obtain a 6 degrees of freedom (DOF) pose of a dynamic object at more than the flicker fusion rate (60 Hz) for geometric alignment of the projection image onto the object. Projector-camera system (ProCams) is one of the suitable system configurations for DPM~\cite{https://doi.org/10.1111/cgf.13387}. Once the geometrical relationship between the camera and the projector is calibrated, the ProCams can align a projection image on the object surface using a 6DOF pose of the object estimated by the camera. The calibration is necessary to determine a correspondence between each point of the object surface and the projector pixel. Moreover, moving the object out of the calibration range causes errors in the pixel correspondence, resulting in significant misalignment of the projected image on the surface.

To solve this problem, a system that uses a beam splitter to make the optical axes of the projector and the camera coaxial has been applied~\cite{1467351}. In the coaxial ProCams, the pixel correspondence between the projection image and the camera image is consistent independently from the pose of the object. Sueishi et al.~\cite{7223330} proposed a coaxial system combined with a saccade mirror to project images on a high-speed moving object. Bermano et al.~\cite{https://doi.org/10.1111/cgf.13128} proposed a markerless DPM on human faces using a coaxial system.

However, in reality, there are a couple of technical difficulties in the implementation of a coaxial system. It requires tedious manual adjustments to perfectly align the optical axes at a sub-pixel accuracy. However, the optical axes of the devices are not known a priori. It is also hard to maintain the coaxial state due to vibration in long-term usage. Even a slight alignment error causes a shift in the pixel correspondence, resulting in significant misalignment of the projection image.

\begin{figure}[t]
  \centering
  \includegraphics[width=0.7\hsize]{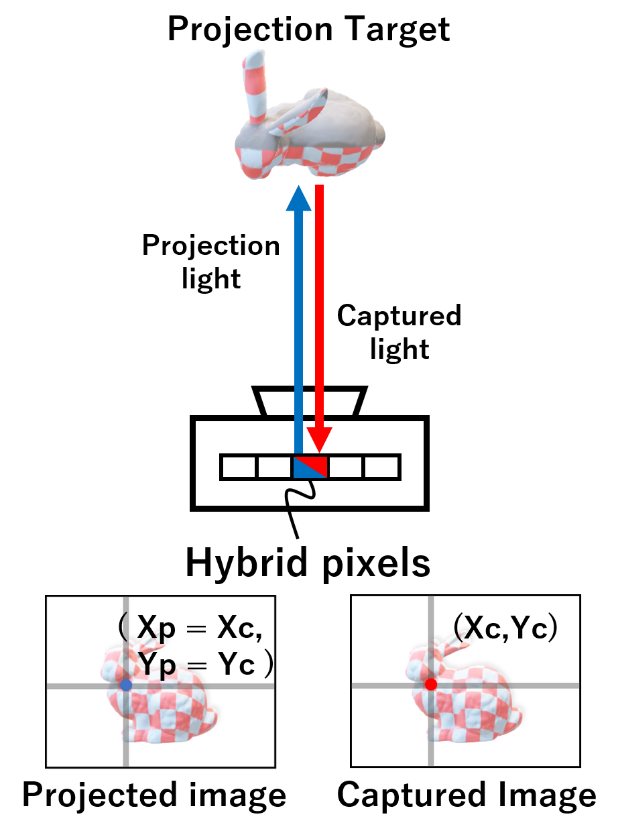}
  \caption{Concept of the proposed ProCams.}
	\label{fig:1}
\end{figure}

For these reasons, in reality, there is no silver bullet to obtain accurate pixel correspondence in DPM using conventional ProCams. In this paper, we propose a novel ProCams in which hybrid pixels have both projecting and capturing capabilities (\autoref{fig:1}). We apply a special display device called bidirectional organic light emitting diode (Bidirectional OLED) in which each pixel consists of RGBW display pixels and a photo sensor. We attach a lens system to the Bidirectional OLED so that the lens is used as a projector lens as well as a camera lens in our ProCams. Owing to the integrated projecting and capturing property, our proposed system provides an entirely consistent pixel correspondence. A large aperture is preferable for a projector to increase the brightness of projected images, and thus, our system applies a large aperture lens system. However, this leads to a narrow depth-of-field (DOF) in both capturing and projection which share the same lens system. We overcome this problem by devising an auto-focus technique with adaptive intrinsic parameters. Finally, we demonstrate the availability of the proposed system in DPM through a projection experiment on a dynamic object.

\section{Proposed Method}

This section describes our proposed ProCams, which employs Bidirectional OLED to make each pixel have both projecting and capturing capabilities. Then, we introduce a method for solving narrow DOF problem caused by that we use the same lens system for both projection and capturing.

\subsection{System Configuration}

\begin{figure}[t]
  \centering
  \includegraphics[width=0.98\hsize]{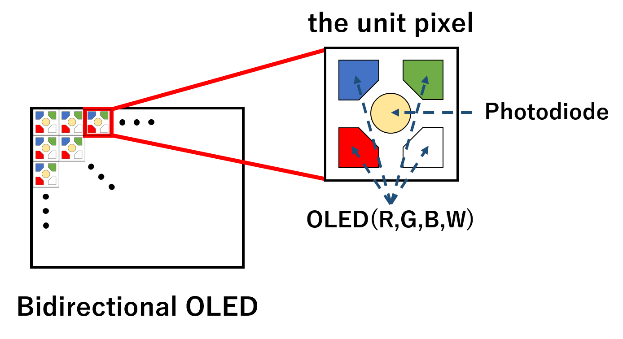}
  \caption{Pixel of Bidirectional OLED.}
	\label{fig:2}
\end{figure}

\begin{figure}[t]
  \centering
  \includegraphics[width=0.98\hsize]{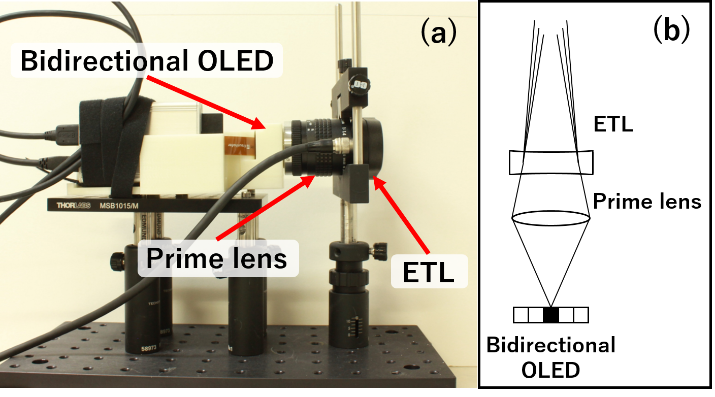}
  \caption{(a) Implemented proposed ProCams. (b) Above image of the Implemented ProCams.}
	\label{fig:3}
\end{figure}

\begin{figure}[t]
  \centering
  \includegraphics[width=0.98\hsize]{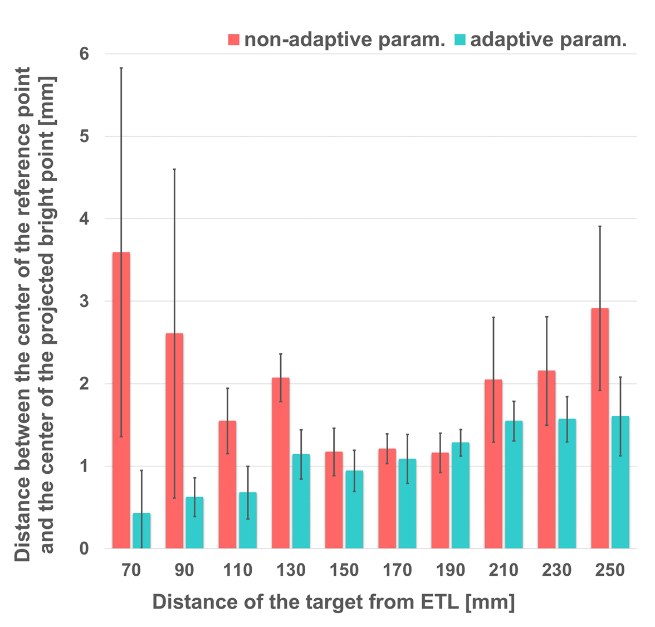}
  \caption{The average distance between the center of the reference point and the center of the projected bright point (Bars are standard deviation).}
	\label{fig:4}
\end{figure}

\begin{figure}[t]
  \centering
  \includegraphics[width=0.98\hsize]{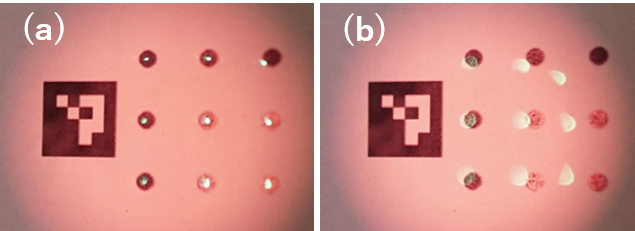}
  \caption{(a) With adaptive intrinsic parameters. (b) With non-adaptive intrinsic parameters (The board was placed at 70 mm from the ETL).}
	\label{fig:5}
\end{figure}

\begin{figure*}[t]
  \centering
  \includegraphics[width=0.75\hsize]{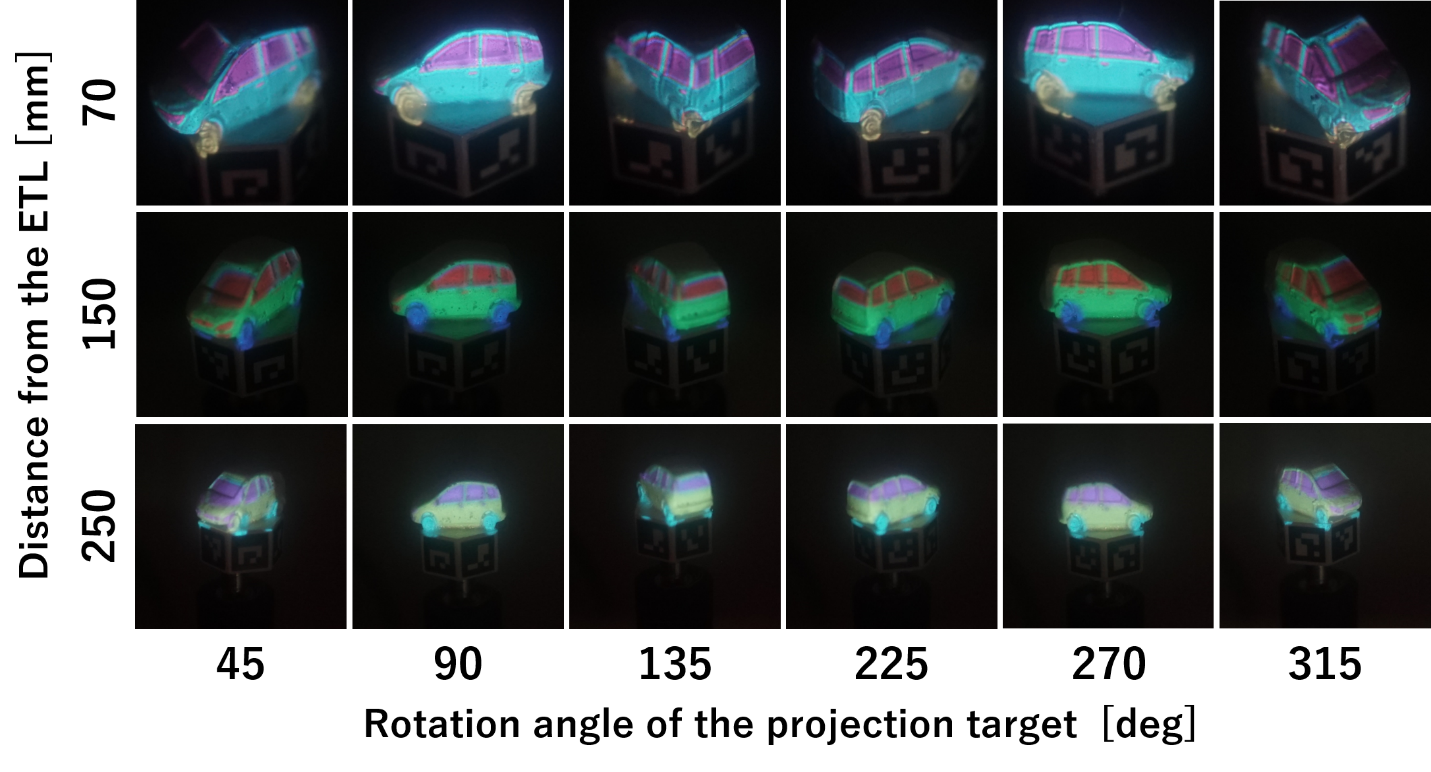}
  \caption{The projection target at distances of 70 mm, 150 mm, and 250 mm from the ETL when projected images onto it using the proposed system.}
	\label{fig:6}
\end{figure*}

Bidirectional OLED is a device that integrates optical sensing functions into the display surface. Each unit pixel consists of one photodiode located in the center of four OLED (\autoref{fig:2}). 

In this study, we propose a ProCams in which projection and capturing share the same image plane using this Bidirectional OLED. We configure our proposed ProCams by combining a Bidirectional OLED device and a lens system. Our system is possible to project a displayed image on the Bidirectional OLED into real space. The system can also acquire real-world appearance as a digital image. In our ProCams, the projector and camera coordinates are completely identical. Therefore, our system does not suffer from the misalignment of projected images on dynamic objects caused by an inaccurate calibration between a projector and camera.

\subsection{Auto-Focus with Adaptive Intrinsic Parameters}

To increase the brightness of the projection image, the aperture of the projector is designed to be large, and thus, the projector's DOF is generally shallow. In the proposed system, the projector and camera share the same lens system. Thus, the camera's DOF also becomes shallow to increase the brightness of the projector. When the object is outside the DOF of the ProCams, the defocus blur occurs in both the captured and projected images. The accuracy of estimating the 6DOF pose of the object worsens because of defocus blur of the captured image. The user's immersion decreases because of defocus blur of the projected image.

We propose an auto-focus system that extends the space in which projection images can be properly projected onto a dynamic object. We apply an electrically focus tunable lens (ETL), which can rapidly change the focal length and has a relatively large aperture compared to other focus tunable mechanisms. Drive current applied to the ETL is computed from the estimated distance from the ETL to the object.

The problem in this auto-focus system is the change in the intrinsic parameters of ProCams as the focal length of the ETL changes. The intrinsic parameters of the ProCams are necessary for the estimation of the 6DOF pose of a projection object. If the incorrect intrinsic parameters are used, the accuracy of the pose estimation decreases. We propose to adjust the intrinsic parameters as the ETL drive current changes. Specifically, in offline, we calibrate intrinsic parameters for different optical powers of the ETL. In run-time, we adjust the intrinsic parameters dynamically according to the ETL's optical power. We obtain intrinsic parameters for uncalibrated optical powers by interpolating the calibrated data. The proposed adaptive intrinsic parameters allow us to estimate the 6DOF pose of a target object without degrading the accuracy while changing the ETL's optical power.

\section{Experiment and Result}

We implemented a proof-of-concept ProCams prototype and conducted an evaluation experiment to validate the proposed auto-focus technique. We then demonstrated a DPM using our system. We conducted these experiments in a darkroom using the system shown in \autoref{fig:3}.

\subsection{Experimental Setup}

Our ProCams prototype consisted of a Bidirectional OLED (Fraunhofer FEP EBCW1020), a prime lens (focal length 12.5 mm), and an ETL (Optotune EL-16-40-TC-20D) (\autoref{fig:3}). In the following experiments, the focal length of the prime lens was fixed so that a projected image is focused at 170 mm from the ETL when the optical power of the ETL was 0 D. In the subsequent experiments, we placed target surfaces between 70 mm and 250 mm from the ETL.

We calibrated the ProCams intrinsic parameters at 10 locations from 70 mm to 250 mm from the ETL in increments of 20 mm. Specifically, we adjusted the drive current to focus at each location where we calibrated the ProCams using Zhang's method~\cite{888718}.

\subsection{Evaluation of Adaptive Intrinsic Parameter}

We conducted an evaluation of our adaptive intrinsic parameter technique. We evaluated the misalignment between reference points printed on a projection target and projected results which were intended to be overlaid on the references.

We used a flat board on which a rectangular augmented reality (AR) marker and reference points were printed. The board was captured with the proposed ProCams to estimate the pose from the AR marker and project a bright spot on each reference point. We placed the board at 10 locations at 70 to 250 mm distance from the ETL in 20 mm increments for projection. An external camera captured the projected results. We prepared two experimental conditions: with adaptive and non-adaptive intrinsic parameters. When intrinsic parameters were not adaptive, we fixed the intrinsic parameter which was obtained when the ETL focal length was adjusted to focus on a target located at 150 mm from the ETL.

\autoref{fig:4} shows the average distance between the center of the reference point and the center of the projected bright point on the board. The results show that adaptive intrinsic parameters according to the optical power of ETL is effective for the geometric alignment of projected images. When intrinsic parameters were not adaptive, errors in the ETL drive current arose because of the reduced accuracy of the AR marker position estimation. This caused a defocus blur in both captured and projected images. \autoref{fig:5} compares projected results with adaptive and non-adaptive intrinsic parameters when the board was placed at 70 mm from the ETL. This figure shows that the projected image was blurred with non-adaptive intrinsic parameters. Therefore, we confirmed that our adaptive intrinsic parameters effectively improved the focusing accuracy of the projected and captured images and consequently, the pose estimation accuracy.

\subsection{Projection onto a Dynamic Object}

We demonstrated the availability of DPM using the proposed system. We evaluated the results of the projected images on a moving object.

The projection target consisted of a car-shaped object and a regular hexagonal prism with AR markers on each side. Each AR marker was 13mm x 13mm in size. The size of the car-shaped object's circumscribed rectangle was 35 mm long x 22 mm wide x 18 mm high. The moving range of the projection target was set from 70 mm to 250 mm away from the ETL. We manually moved the projection target on a linear translation stage. To prevent crosstalk between the captured light and the projected light, we illuminated the target with infrared light.

The proposed system firstly estimated the 6DOF pose of the target from the captured infrared image. Then, the ETL drive current was altered to auto-focus on the target. The projection image was also generated based on the estimated 6DOF pose of the target. Finally, the system projected an image corrected by the Winner filter. This was to compensate for the defocus blur caused by chromatic aberration in infrared and visible light. To visualize the target's movement, the proposed system changed projection texture colors. Specifically, the body color was blue, green, and yellow when the car object located from 70 mm to 130 mm, from 130 mm to 190 mm, and from 190 mm to 250 mm away from the ETL, respectively.

\autoref{fig:6} shows the projection results taken by an external camera. This figure shows the projected results when the object was 70 mm, 150 mm, and 250 mm away from the ETL. From this result, it was confirmed that the projected image was aligned on the target. It was also confirmed that the color of the projected texture changed according to the distance of the target from the ETL. The movable range of the target was 70 to 250 mm from the ETL. The processing time for the whole process was 194 ms. The experimental results show that the proposed system is applicable to project images onto a slowly moving object.

\section{Conclusion}

In this study, we proposed a new conceptual configuration of ProCams, in which hybrid pixels have both projection and capturing capabilities. This was achieved by using a Bidirectional OLED, whose pixel consists of a photodiode and full color display pixels. We demonstrated the availability of the proposed system for DPM through the experiments shown in Sec. 3. The results showed that the proposed system was promising as one of implementations of DPM. As a future work, we try to confirm the performance of DPM with quickly moving objects.

\acknowledgments{
This work was supported by JSPS KAKENHI grant number JP17H04691.}

\bibliographystyle{abbrv-doi}

\bibliography{template}
\end{document}